\documentclass[sigconf]{acmart}
\AtBeginDocument{%
  }

\setcopyright{acmlicensed}

\copyrightyear{2026}
\acmYear{2026}
\setcopyright{cc}
\setcctype{by}
\acmConference[MM '26]{Proceedings of the 34th ACM International Conference on Multimedia}{November 10--14, 2026}{Rio de Janeiro, Brazil}
\acmBooktitle{Proceedings of the 34th ACM International Conference on Multimedia (MM '26), November 10--14, 2026, Rio de Janeiro, Brazil}
\acmDOI{10.1145/3767308.3836527}
\acmISBN{979-8-4007-2213-4/2026/11}

\usepackage{multirow}
\usepackage[table]{xcolor}
\usepackage{enumitem}
\usepackage[capitalize,noabbrev]{cleveref}
\usepackage{xspace}
\usepackage{pifont}
\usepackage[utf8]{inputenc}
\usepackage{balance}

\newcommand{\ours}[0]{\textsc{CharTool}\xspace}
\newcommand{\data}[0]{\textsc{DuoChart}\xspace}

\definecolor{rowgray}{gray}{0.95} %
\definecolor{rowpurple}{HTML}{F4F0F8} %
\definecolor{deltared}{HTML}{B01919} 

\newcommand{\dplus}[1]{\textcolor{deltared}{\textbf{+#1}}}

\begin{document}

\title{CharTool: Tool-Integrated Visual Reasoning for Chart Understanding}

\author{Situo Zhang}
\authornote{Both authors contributed equally to this research.}
\affiliation{%
  \institution{X-LANCE Lab, Shanghai Jiao Tong University}
  \city{Shanghai}
  \country{China}
}
\email{situozhang@sjtu.edu.cn}

\author{Yifan Zhang}
\authornotemark[1]
\affiliation{%
  \institution{X-LANCE Lab, Shanghai Jiao Tong University}
  \city{Shanghai}
  \country{China}
}
\email{zhang-yifan@sjtu.edu.cn}

\author{Zichen Zhu}
\affiliation{%
  \institution{X-LANCE Lab, Shanghai Jiao Tong University}
  \city{Shanghai}
  \country{China}
}
\email{jameszhuthethird@sjtu.edu.cn}

\author{Da Ma}
\affiliation{%
  \institution{X-LANCE Lab, Shanghai Jiao Tong University}
  \city{Shanghai}
  \country{China}
}
\email{mada123@sjtu.edu.cn}

\author{Lei Pan}
\affiliation{%
  \institution{East China Normal University}
  \city{Shanghai}
  \country{China}
}
\email{leipan@stu.ecnu.edu.cn}

\author{Danyang Zhang}
\affiliation{%
  \institution{X-LANCE Lab, Shanghai Jiao Tong University}
  \city{Shanghai}
  \country{China}
}
\email{zhang-dy20@sjtu.edu.cn}

\author{Zihan Zhao}
\affiliation{%
  \institution{X-LANCE Lab, Shanghai Jiao Tong University}
  \city{Shanghai}
  \country{China}
}
\email{zhao\_mengxin@sjtu.edu.cn}

\author{Lu Chen}
\affiliation{%
  \institution{X-LANCE Lab, Shanghai Jiao Tong University}
  \institution{Shanghai Innovation Institution}
  \city{Shanghai}
  \country{China}
}
\affiliation{%
  \institution{Jiangsu Key Lab of Language Computing}
  \city{Suzhou}
  \country{China}
}
\correspondingauthor
\email{chenlusz@sjtu.edu.cn}

\author{Kai Yu}
\affiliation{%
  \institution{X-LANCE Lab, Shanghai Jiao Tong University}
  \city{Shanghai}
  \country{China}
}
\affiliation{%
  \institution{Jiangsu Key Lab of Language Computing}
  \city{Suzhou}
  \country{China}
}
\correspondingauthor
\email{kai.yu@sjtu.edu.cn}

\renewcommand{\shortauthors}{Situo Zhang et al.}

\begin{abstract}
  Charts are ubiquitous in scientific and financial literature for presenting structured data. However, chart reasoning remains challenging for multimodal large language models (MLLMs) due to the lack of high-quality training data, as well as the need for fine-grained visual grounding and precise numerical computation. To address these challenges, we first propose \data, a scalable dual-source data pipeline that combines synthesized charts with real-world charts to construct diverse, high-quality chart training data. We then introduce \ours, which equips MLLMs with external tools, including image cropping for localized visual perception and code-based computation for accurate numerical reasoning. Through agentic reinforcement learning on \data, \ours learns tool-integrated reasoning grounded in chart content. Extensive experiments on six chart benchmarks show that our method consistently improves over strong MLLM baselines across model scales. Notably, \ours-7B outperforms the base model by \textbf{+8.0\%} on CharXiv (Reasoning) and \textbf{+9.78\%} on ChartQAPro, while achieving competitive performance with substantially larger or proprietary models. Moreover, \ours demonstrates positive generalization to out-of-domain visual math reasoning benchmarks. Our data and models are available at \url{https://github.com/OpenDFM/CharTool}.
  
\end{abstract}




\maketitle

\begin{figure*}
\centering
    \includegraphics[width=\textwidth]{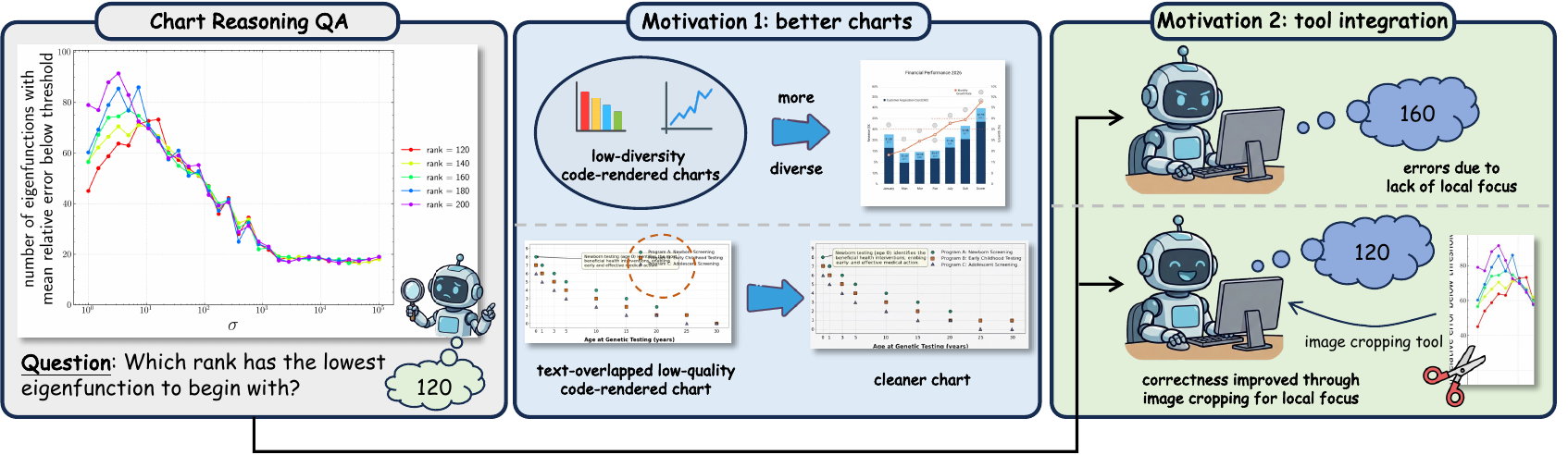}
\caption{
Motivation for our method.
(Left) Chart reasoning requires fine-grained visual perception and numerical reasoning.
(Middle) Existing synthetic chart datasets often lack diversity and visual quality.
(Right) Purely text-based reasoning leads to errors on complex layouts, while explicit tool grounding (e.g., image cropping) enables more accurate, localized analysis.
}
\Description{An introductory figure explaining our motivation.}
    \label{fig:intro}
\end{figure*}

\section{Introduction}
Charts serve as a fundamental medium for presenting structured data in scientific papers and financial reports, combining textual annotations with structured visual elements to convey quantitative relationships. Although recent multimodal large language models (MLLMs) have achieved strong performance in general visual understanding and reasoning~\cite{gpt4, llava1, gemini25, qwen3vl, vision-r1, zhu2025multi}, they continue to struggle with chart reasoning, which requires fine-grained visual perception as well as precise relational and numerical reasoning over densely structured information~\cite{chartqa, charxiv, chartbench}.

To tackle chart reasoning, existing research has primarily focused on either scaling up training data through synthesis or improving models' reasoning capabilities for chart question answering~\citep{matcha, chartllama, mmc, chartassistant}. However, current works suffer from two key limitations. 1) \textbf{Existing synthesized chart datasets often lack sufficient diversity and complexity}~\citep{charxiv}. Early approaches rely on fixed templates and cover only a limited set of chart types~\citep{chartllama, opencqa}. More recent code-driven pipelines are more scalable, but they still struggle to capture real-world, long-tail chart patterns and often produce degraded layouts (e.g., overlapping text) as chart complexity increases~\citep{reachqa, ECD}. 2) \textbf{Current chart reasoning methods are constrained by the limitations of chain-of-thought text-based reasoning}. Unlike general visual question answering, chart reasoning requires fine-grained visual and numerical analysis to interpret complex elements such as overlapping data points, intersecting trend lines, and dense numerical annotations, while simultaneously modeling their spatial arrangements and semantic meanings. However, existing methods typically generate answers either directly or through textual reasoning~\citep{chartr1,bigchartsr1,cot} (\Cref{fig:intro}-Right), focusing on textual logical inference rather than grounded visual processing. As a result, they are often reliant on language priors, making them prone to hallucination~\citep{hallucination} and limiting their ability to perform precise numerical computation or capture complex relationships encoded in charts~\citep{deepeyes,thyme}.

To address these limitations and further improve the chart reasoning capabilities of MLLMs, we propose \data, a \textit{dual-source} chart synthesis pipeline~(\S\ref{sec:chart-image}). The two data sources offer complementary strengths: real-world charts from scientific literature provide broad coverage and high visual quality, while code-driven synthetic charts provide precise underlying data for answer generation and verification. To combine these advantages, we sample diverse and fine-grained specifications, including chart types, layouts, and difficulty levels, to generate varied chart-rendering code. We further incorporate high-quality charts from scientific literature to better capture real-world chart distributions. In addition, we design a dedicated question-answer synthesis pipeline with multi-stage checking and scoring to construct high-quality, challenging QA training data grounded in the generated charts~(\S\ref{sec:qa-generation}). Using this pipeline, we construct \data-100k, which contains high-quality charts and 100k challenging QA samples.

Second, to address the limitations of text-based reasoning for fine-grained visual and numerical chart understanding, we introduce \mbox{\ours}, which equips MLLMs with tool-integrated reasoning capabilities~(\S\ref{sec:tir}). In particular, \mbox{\ours} introduces an \textbf{image-cropping tool} for capturing localized visual cues, together with \textbf{code-based computation tools} for accurate numerical calculation. We train the model using end-to-end reinforcement learning~\citep{searchr1,torl,toolrl,r1searcher} within an isolated sandbox environment, with a tool-aware reward to guide effective tool usage. This enables the model to learn how to invoke tools, generate executable tool actions, and incorporate returned observations into subsequent reasoning, thereby supporting effective multi-step, tool-integrated reasoning.

Experimental results on six chart reasoning benchmarks show that \mbox{\ours}, trained on \data and equipped with tool-grounded reasoning, consistently improves performance across different model scales~(\S\ref{sec:main-results}). Notably, on the challenging real-world benchmark CharXiv~\citep{charxiv}, \mbox{\ours}-7B achieves an accuracy of 50.5\%, corresponding to an \textbf{8.0\% absolute improvement}.
Moreover, \mbox{\ours}-7B yields a \textbf{9.78\% absolute improvement} on ChartQAPro~\citep{chartqapro}.
Furthermore, \mbox{\ours} demonstrates strong generalization to three out-of-domain visual mathematical reasoning benchmarks~(\S\ref{sec:ood-result}), indicating that the tool-integrated reasoning ability learned from charts can transfer beyond the chart domain.

Our main contributions are summarized as follows:
\begin{itemize}[left=3pt]
    \item We develop a multi-agent chart synthesis and question–answer generation pipeline that produces diverse, high-quality charts and challenging QA pairs.
    \item We enable MLLMs to leverage external tools for precise numerical computation and fine-grained visual reasoning, with agentic reinforcement learning used to train effective tool-integrated chart reasoning.
    \item We demonstrate the effectiveness of our approach across six chart reasoning benchmarks, achieving substantial gains on challenging real-world datasets and showing strong generalization to out-of-domain visual mathematical reasoning tasks.
\end{itemize}

\begin{figure*}[t]
    \centering
    \includegraphics[width=0.85\linewidth]{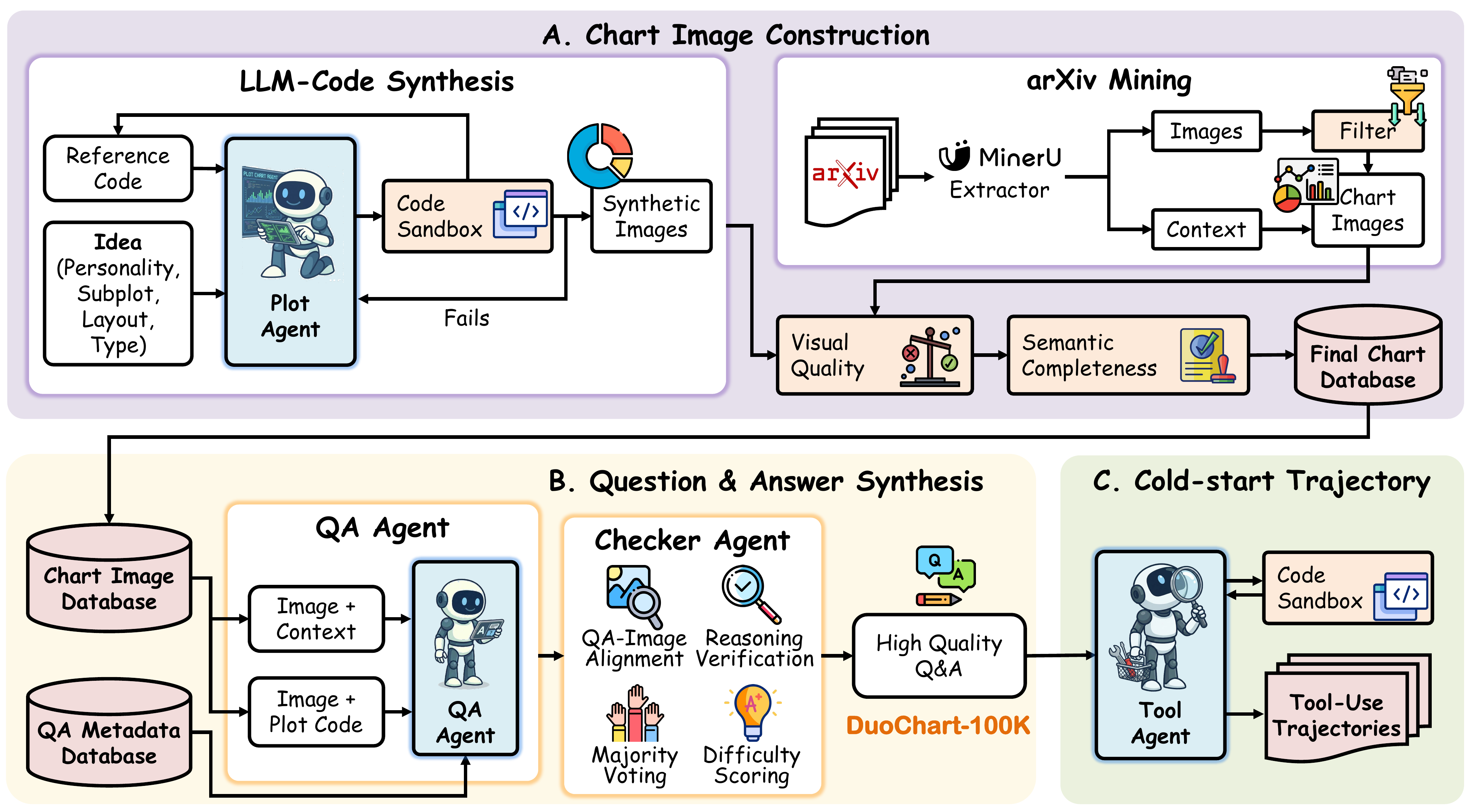}
    \caption{Data synthesis pipeline of \data. (A) Dual-source chart image construction. (B) High-quality QA generation and validation. (C) Cold-start trajectory synthesis.}
    \label{fig:data-pipeline}
\end{figure*}

\section{Method}
In this section, we first describe \data, our multi-agent chart data synthesis pipeline, with a focus on chart image construction and question-answer generation. We then introduce \ours, our tool-integrated reasoning framework for chart understanding.

\subsection{Multi-Agent Chart Data Synthesis}
\subsubsection{Hybrid Chart Image Construction}
\label{sec:chart-image}

We observe a fundamental trade-off between the two data sources. Real-world charts collected from scientific literature exhibit broad distributions and generally high visual quality, but they do not provide accessible underlying data values, making it difficult to construct complex reasoning questions and derive verifiable ground-truth answers. In contrast, code-driven synthetic charts provide precise underlying data for answer generation and verification, yet often fall short in visual quality and diversity compared with real-world charts.

To leverage their complementary strengths, we construct a hybrid chart image corpus from two sources, as illustrated in \Cref{fig:data-pipeline}-A: (i) code-driven synthetic charts for controllable diversity and precise labels, and (ii) real-world charts mined from scientific papers for realistic visual complexity and broader chart distributions. By combining both sources, our corpus balances controllability, annotation quality, and realism, providing a stronger foundation for downstream chart QA generation.

\paragraph{\textbf{Code-Based Synthetic Chart Generation.}}
To ensure consistency with real-world chart distributions, we curate a \emph{Reference Code} library by reproducing real-world charts with MLLM-generated Python code and manually verifying alignment with the originals. This process yields a set of 112 reference charts, including both single-plot and multi-subplot layouts. To promote topic diversity, we adopt a persona-based sampling strategy~\cite{personas-dataset}. Conditioned on sampled attributes such as subplot count, layout, and chart type, a Plot Agent generates plotting code, which is executed in a sandbox environment with iterative correction and refinement until valid charts are rendered. This pipeline enables scalable synthesis of diverse charts while preserving realistic structure and accurate underlying data.

\paragraph{\textbf{Mining Real-World Charts from Papers.}}
To expose the model to realistic chart layouts and visual diversity, we mine charts from arXiv papers\footnote{\url{https://arxiv.org/}. We collect papers from October 2024 to October 2025, with no overlap with CharXiv~\cite{charxiv}.} spanning eight fields. We use MinerU~\cite{mineru25} to extract figure images together with their associated textual context. Semantic filtering is then applied using an MLLM to retain only valid data visualizations and remove non-chart figures or overly cluttered elements. This real-world collection complements the synthetic charts by introducing visual complexity, layout variation, and domain diversity.

\paragraph{\textbf{Image Quality Filtering.}}
We apply automatic filtering to ensure both visual usability and semantic completeness. An MLLM-based judge scores each candidate chart along two axes: (1) \textbf{visual quality}, which penalizes severe overlap, misalignment, and unreadable text; and (2) \textbf{semantic completeness}, which checks whether essential elements such as axes, legends, and labels are present for self-contained interpretation. Charts failing either criterion are discarded, yielding a final high-quality chart image database suitable for fine-grained chart reasoning.

\begin{figure*}[t]
    \centering
    \includegraphics[width=\linewidth]{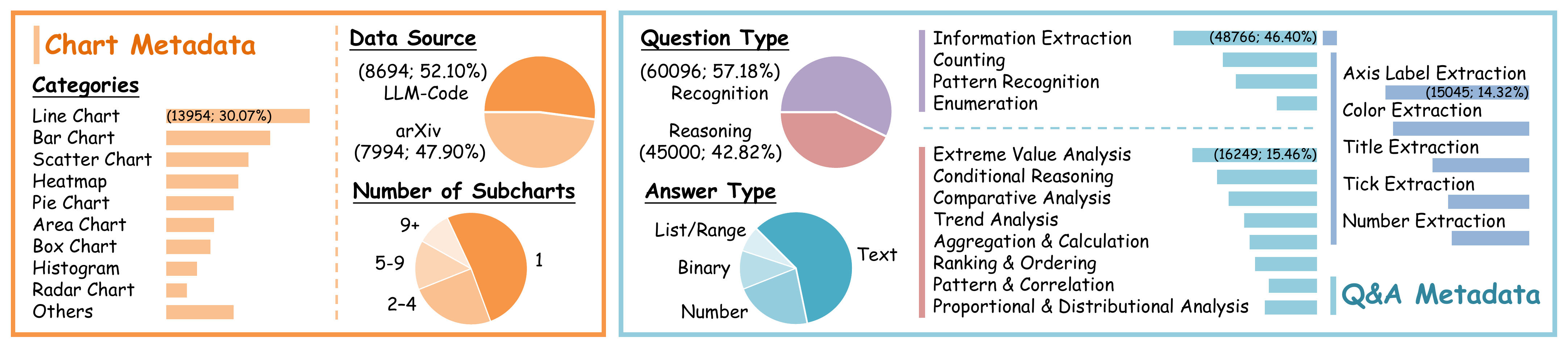}
    \caption{Data statistics of the charts (Left) and QAs (Right) in \data.}
    \label{fig:data-stat}
\end{figure*}

\subsubsection{Reliability-Driven QA Generation}
\label{sec:qa-generation}

\paragraph{\textbf{Question-Answer Synthesis.}}
As shown in \Cref{fig:data-pipeline}-B, given a chart image and its associated context, the multimodal QA Agent generates candidate question-answer pairs spanning two complementary categories: \textbf{recognition} questions for element identification and structural extraction, and \textbf{reasoning} questions that require multi-step visual analysis and numerical reasoning. To enable both diversity and complexity, we construct questions by conditioning the QA Agent on fine-grained analytical aspects drawn from a QA metadata pool. These aspects explicitly specify what type of visual or numerical reasoning is required, such as legend interpretation, cross-subplot comparison, trend estimation, or multi-variable aggregation. By sampling across these fine-grained aspects and conditioning the generation process accordingly, the resulting questions cover a broad spectrum of reasoning behaviors with varying difficulty levels.

\paragraph{\textbf{QA Quality Control.}}
Following the generation phase, we apply a Checker Agent to filter hallucinations and retain reliable supervision. Specifically, we (1) enforce \textbf{QA-image alignment} to remove questions that depend on non-visual knowledge or unsupported claims; (2) perform \textbf{reasoning verification} to ensure the answer is correct and the reasoning is consistent with the chart evidence; (3) use \textbf{majority voting} to discard ambiguous cases with low agreement; and (4) assign \textbf{difficulty scores} and keep questions that require non-trivial visual and numerical reasoning. This process yields our high-quality QA dataset \data-100k. We present the statistics of our data in \Cref{fig:data-stat}.

\subsubsection{Cold-Start Trajectory Synthesis}

To initialize tool-use behavior, we construct a cold-start dataset of executable trajectories~(\Cref{fig:data-pipeline}-C) sampled from \data. For each chart question, a Tool Agent is prompted to solve the problem step by step, invoking image cropping or computation tools as needed. Each tool call is executed in a code sandbox, with outputs appended to the context for subsequent reasoning. The full sequence of reasoning steps, tool calls, and observations forms a trajectory. We retain only those with successful execution and correct final answers, resulting in about 9k reliable demonstrations for cold-start supervised fine-tuning.

\subsection{Tool-Integrated Chart Reasoning}
\label{sec:tir}
We introduce tool-integrated reasoning~(TIR) for chart understanding, where an MLLM alternates between internal reasoning and external tool invocation. Given an input chart image $\mathcal{I}$ and a question $q$, the model follows a policy $\pi_\theta$ that iteratively generates reasoning tokens and selects tool actions based on the current context. This interaction continues until the model outputs a final answer. Formally, the model generates a trajectory
\begin{equation}
    \tau = (\mathcal{I}, q, r_1, a_1, o_1, \dots, r_n, a_n, o_n, r_{n+1}, y),
\end{equation}
where $r_i$ denotes the reasoning tokens at step $i$, $a_i$ represents the selected tool action (e.g., image cropping or computation), $o_i$ is the observation returned by the tool execution, and $y$ is the final predicted answer. Observations $o_i$ may consist of text outputs or processed visual inputs, which are appended to the model context for subsequent reasoning.

\subsubsection{Reinforcement Learning for TIR}
\label{sec:tool-rl}
Starting from the cold-start model, we further optimize the policy $\pi_\theta$ with reinforcement learning to improve tool usage and multi-step reasoning. The objective is to maximize the expected reward over trajectories:
\begin{equation}
    \mathcal{J}(\theta) = \mathbb{E}_{\tau \sim \pi_\theta} \left[ R(\tau) \right],  
\end{equation}
where $R(\tau)$ evaluates the quality of the final answer $y$ with respect to the ground-truth answer $y^*$. In our implementation, the reward is defined at the trajectory level based on answer correctness, while intermediate reasoning and tool usage are optimized implicitly through policy gradients.

\subsubsection{Policy Optimization with GRPO}
To optimize the policy, we adopt Group Relative Policy Optimization (GRPO)~\cite{grpo}. Let $\mathcal{M} \subseteq \{1, \dots, L\}$ denote the set of token indices corresponding to reasoning steps, tool calls, and final answer tokens that are not masked during RL training. Tokens outside $\mathcal{M}$ (e.g., prompt or observation tokens) are masked for training.

At each update step, we sample a group of $G$ trajectories $\{\tau_g\}_{g=1}^G$ from the previous policy $\pi_{\theta_{\text{old}}}$. The GRPO objective is defined as

\begin{equation}
\begin{aligned}
\mathcal{J}_{\text{GRPO}}(\theta)
&=
\mathbb{E}_{\{\tau_g\}_{g=1}^G \sim \pi_{\theta_{\text{old}}}}
\Bigg[
\frac{1}{G}
\sum_{g=1}^G
\frac{1}{\lvert \mathcal{M} \rvert}
\sum_{j \in \mathcal{M}}
\\
\min  \Big(
w_g^j&(\theta) A_g^j,\;
\text{clip}\big(w_g^j(\theta), 1-\epsilon, 1+\epsilon\big) A_g^j
\Big)
\Bigg],
\end{aligned}
\end{equation}
where $
    w_g^j(\theta) =
    \frac{\pi_\theta(x_j \mid x_{<j})}
         {\pi_{\theta_{\text{old}}}(x_j \mid x_{<j})}
$; $\epsilon$ is the clipping threshold; and $A_g^j$ denotes the token-level advantage computed by normalizing the trajectory-level reward within each group:

\begin{equation}
    A_g = \frac{r_g-\text{mean}(\{R_g\}_{g=1}^G)}{\text{std}(\{R_g\}_{g=1}^G)}.
\end{equation}
\paragraph{\textbf{Tools.}} In this work we focus on two tools that are most critical for chart understanding: an \textbf{image cropping} tool for fine-grained visual perception, and a \textbf{code computation} tool for explicit numerical operations such as aggregation and statistical analysis. This design enables the model to seamlessly interleave visual localization and precise computation with the reasoning process.

\paragraph{\textbf{Reward Design.}}
We design the reward function comprising three parts: (1) \textbf{Accuracy reward $R_\text{acc}$}, which evaluates the correctness of the generated response; (2) \textbf{Format reward $R_\text{format}$}, which enforces structural compliance across reasoning and tool-calling templates to ensure reliable parsing; and (3) \textbf{Tool reward $R_\text{tool}$}, which encourages the exploration of tool-assisted trajectories to balance the exploitative tendency of \textbf{$R_{acc}$}. Each of these reward components is implemented as a binary signal (1.0/0.0). The  total reward is defined as:
\begin{equation}
    R(\tau) = R_\text{acc}(\tau) + \lambda_1\cdot R_\text{format}(\tau) + \lambda_2\cdot \mathbb{I}_{R_\text{acc}(\tau) > 0} \cdot R_\text{tool}(\tau),
\end{equation}
where $\mathbb{I}_{R_\text{acc}(\tau) > 0}$ is an indicator function that is 1 if the answer is correct and 0 otherwise, ensuring that tool calls are encouraged only when the answer is correct. We empirically set $\lambda_1 = 0.1$ and $\lambda_2 = 0.2$ in experiments.

\section{Experiments}

In this section, we thoroughly evaluate our method from three perspectives. First, we assess the quality of our synthesized dataset \data. Second, we implement \ours and evaluate its performance on a diverse set of chart reasoning benchmarks, as well as out-of-domain visual reasoning tasks. Third, we conduct ablation studies to analyze the impact of key design components.

\subsection{Quality of the Synthesized Dataset}
\label{sec:dataquality}

As shown in \Cref{fig:data-stat}, our dataset covers a wide variety of chart layouts, chart types, and question types. We further evaluate the quality of our synthesized dataset \data through both quantitative and model-based evaluation. Specifically, we report average entropy of chart images and employ GPT-5.2 to judge four key metrics: visual quality, question-image alignment, answer correctness, and reasoning difficulty (further details are provided in supplementary material). For comparison, we benchmark \data against widely adopted synthetic chart datasets, including ReachQA~\cite{reachqa} and ECD~\cite{ECD}.

\begin{figure}[h]
\centering
\includegraphics[width=\linewidth, trim= 0 20 0 0, clip]{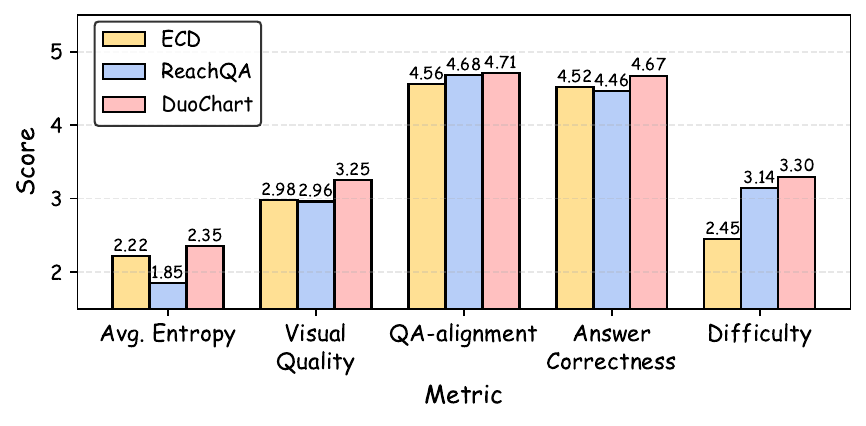}
\caption{Comparison of synthesized dataset quality.}
\label{fig:data-quality}
\end{figure}

As shown in \Cref{fig:data-quality}, \data consistently surpasses both ReachQA and ECD across all evaluation criteria. Notably, it achieves higher image entropy, which indicates the presence of more complex layouts. It also demonstrates superior visual quality and includes a greater proportion of challenging, high-difficulty reasoning questions. These results demonstrate that \data more accurately reflects the complexity and diversity of real-world chart analysis tasks.

\subsection{Experimental Setup}

\paragraph{Setup.}
We implement \ours using the Qwen2.5-VL architecture with both 3B and 7B parameter variants~\cite{qwen25vl}. Training is conducted in two stages: supervised fine-tuning (SFT) on cold-start trajectories, followed by reinforcement learning on \data. Additional training hyperparameters and implementation details are provided in the supplementary material.

\paragraph{Evaluation.}
We conduct a comprehensive evaluation of our models across a diverse set of standard chart benchmarks, categorized into (1) \textbf{Real-world Benchmarks}, comprising charts extracted from real-world sources, including CharXiv~\cite{charxiv}, ChartQAPro~\cite{chartqapro}, and ChartQA~\cite{chartqa}; and (2) \textbf{Synthetic Benchmarks}, containing synthesized images or queries: ChartBench~\cite{chartbench}, ChartX~\cite{chartx}, and ReachQA~\cite{reachqa}. Furthermore, to assess the out-of-domain generalization capabilities of our approach, we include three visual math reasoning benchmarks: MathVista~\cite{mathvista}, WeMath~\cite{wemath}, and MathVerse~\cite{mathverse}. Further evaluation details are available in supplementary material.

\paragraph{Baselines.}
We compare \ours against three classes of models: (1) \textbf{Proprietary and general-purpose MLLMs}, including GPT-4o~\cite{gpt4o}, Claude~3.5 Sonnet~\cite{claude35}, InternVL3-8B~\cite{internvl3}, Qwen3-VL-8B~\cite{qwen3vl}, and the Qwen2.5-VL series~\cite{qwen25vl}; (2) \textbf{Chart-specialized models}, including TinyChart~\cite{tinychart}, ChartGemma~\cite{chartgemma}, ChartMoE~\cite{chartmoe}, and ECD-Qwen2.5-VL-7B~\cite{ECD}; and (3) \textbf{think-with-image models}, including DeepEyes~\cite{deepeyes}, DeepEyes-V2~\cite{deepeyesv2} and Thyme~\cite{thyme}.

\subsection{Main Results}
\label{sec:main-results}

\begin{table*}[t]
\caption{Performance comparison across diverse chart benchmarks. The best result in each column is highlighted in \textbf{bold}, the second-best result is \underline{underlined}, and relative improvements over the base model ($\Delta$) are shown in \textcolor{deltared}{\textbf{red}}.}

\label{tab:main-result}
\centering
\small
\setlength{\tabcolsep}{6.5pt} %
\renewcommand{\arraystretch}{1.0} %
\begin{tabular}{l c cc c c cc c cc c}
\toprule
\multirow{2.5}{*}{\textbf{Method}} & \multirow{2.5}{*}{\textbf{Param.}} & \multicolumn{2}{c}{\textbf{CharXiv}} & \textbf{ChartQA} & \textbf{ChartQAPro} & \multicolumn{2}{c}{\textbf{ChartBench}} & \textbf{ChartX} & \multicolumn{2}{c}{\textbf{ReachQA}} & \multirow{2.5}{*}{\textbf{Avg.}} \\
\cmidrule(lr){3-4} \cmidrule(lr){5-5} \cmidrule(lr){6-6} \cmidrule(lr){7-8} \cmidrule(lr){9-9} \cmidrule(lr){10-11}

& & \textit{Reas.} & \textit{Desc.} & \textit{Avg.} & \textit{Overall} & \textit{NQA} & \textit{Binary} & \textit{QA} & \textit{Reas.} & \textit{Recog.} & \\
\midrule

\rowcolor{rowgray} \multicolumn{12}{c}{\textit{\textbf{Proprietary and General Model}}} \\
GPT-4o & -- & 47.10 & \underline{84.45} & 85.70 & 37.67 & 52.88 & 81.03 & 58.33 & 39.70 & 66.80 & 61.52 \\
\rowcolor{rowpurple} Claude Sonnet 3.5 & -- & \textbf{60.20} & 84.30 & \textbf{90.80} & 43.58 & 48.29 & 76.72 & 42.71 & \textbf{51.70} & \textbf{74.30} & 63.62 \\
InternVL3-8B & 8B & 39.50 & 75.33 & 82.72 & 37.78 & 45.95 & 84.18 & 54.60 & 30.50 & 59.40 & 56.66 \\
\rowcolor{rowpurple} Qwen3-VL-8B & 8B & 46.40 & 83.00 & 84.44 & 42.65 & 57.86 & 76.83 & 59.72 & \underline{47.70} & 70.70 & 63.26 \\
Qwen2.5-VL-72B & 72B & 49.70 & \textbf{87.40} & \underline{89.50} & \underline{47.04} & \textbf{59.62} & 84.45 & 60.68 & 43.90 & 71.90 & \underline{66.02} \\
\rowcolor{rowpurple} Qwen2.5-VL-3B & 3B & 31.30 & 58.60 & 76.64 & 30.28 & 48.10 & 77.26 & 53.91 & 26.80 & 59.20 & 51.34 \\
Qwen2.5-VL-7B & 7B & 42.50 & 73.60 & 84.72 & 39.84 & 57.67 & 87.58 & \underline{60.85} & 37.70 & 70.50 & 61.66 \\

\midrule

\rowcolor{rowgray} \multicolumn{12}{c}{\textit{\textbf{Chart-Specialized Model}}} \\
TinyChart & 3B & 12.20 & 16.12 & 70.00 & 13.25 & 29.67 & 47.87 & 32.29 & 6.20 & 21.10 & 27.63 \\
\rowcolor{rowpurple} ChartGemma & 3B & 21.60 & 15.88 & 65.12 & 6.84 & 29.86 & 85.94 & 27.78 & 7.30 & 26.90 & 31.91 \\
ChartMOE & 8B & 28.30 & 47.42 & 69.68 & 29.10 & 31.48 & 62.86 & 34.38 & 11.50 & 37.30 & 39.11 \\
\rowcolor{rowpurple} ECD Qwen2.5-VL-7B & 7B & 41.80 & 75.08 & 81.44 & 22.50 & 53.90 & 75.52 & 60.76 & 36.20 & 69.50 & 57.41 \\

\midrule

\rowcolor{rowgray} \multicolumn{12}{c}{\textit{\textbf{Think-with-Image Model}}} \\
 Thyme & 7B & 42.50 & 65.33 & 82.24 & 46.55 & 53.10 & \underline{90.52} & 59.11 & 38.80 & 66.90 & 60.56 \\
\rowcolor{rowpurple} Deepeyes & 7B & 41.40 & 64.95 & 79.76 & 45.55 & 55.95 & 87.63 & 58.42 & 38.50 & 68.30 & 60.05 \\
Deepeyes-V2 & 7B & 48.90 & 78.60 & 83.60 & 47.88 & 57.76 & 71.05 & 58.16  & 43.10 & 68.00 & 61.89 \\

\addlinespace
\midrule

\rowcolor{rowgray} \multicolumn{12}{c}{\textit{\textbf{\ours (Ours)}}} \\
\textbf{\ours-3B} & \textbf{3B} & 41.40 & 74.75 & 81.68 & 42.54 & 51.71 & 84.51 & 55.38 & 36.70 & 61.40 & 58.90 \\
 $\Delta$ (\textit{vs} Qwen2.5-VL-3B) & & \dplus{10.10} & \dplus{16.15} & \dplus{5.04} & \dplus{12.26} & \dplus{3.61} & \dplus{7.25} & \dplus{1.47} & \dplus{9.90} & \dplus{2.20} & \dplus{7.55} \\
\midrule
\addlinespace
\textbf{\ours-7B} & \textbf{7B} & \underline{50.50} & 80.30 & 86.76 & \textbf{49.62} & \underline{58.95} & \textbf{91.58} & \textbf{62.33} & 46.30 & \underline{72.30} & \textbf{66.52} \\
 $\Delta$ (\textit{vs} Qwen2.5-VL-7B) & & \dplus{8.00} & \dplus{6.70} & \dplus{2.04} & \dplus{9.78} & \dplus{1.28} & \dplus{4.00} & \dplus{1.48} & \dplus{8.60} & \dplus{1.80} & \dplus{4.85} \\
\bottomrule
\end{tabular}
\end{table*}

To thoroughly evaluate \ours, we present the performance on benchmarks covering diverse chart types in \Cref{tab:main-result}. \ours demonstrates consistent and notable gains across diverse benchmarks and model sizes. Based on the quantitative results, we make the following key observations:

\paragraph{\ours yields notable improvements over baselines across all benchmarks.}
Compared to the Qwen2.5-VL backbone models, \ours achieves substantial improvements across all evaluation metrics. On the challenging real-world benchmark CharXiv, \ours-3B and \ours-7B achieve absolute improvements of \textbf{10.1\%} and \textbf{8.0\%} in reasoning accuracy, respectively. Compared to previous chart-specialized models, \ours demonstrates superior performance on real-world datasets and achieves significant gains in average accuracy. These consistent improvements confirm that our tool-integrated approach effectively enhances the model's ability to interpret and reason over chart data.

\paragraph{Comparable Performance with Proprietary and Large-Scale Models.}
Despite using only 7B parameters, \ours-7B achieves performance comparable to advanced proprietary and large-scale open-source models. As shown in \Cref{tab:main-result}, \ours-7B attains an average score of \textbf{66.52}, outperforming GPT-4o (61.52) and Claude~3.5 Sonnet (63.62), and the much larger Qwen2.5-VL-72B (66.02). These results suggest that our tool-integrated reasoning framework provides a more parameter-efficient approach to chart reasoning than scaling model size alone.

\paragraph{Comparison with Think-with-Image Models.}
\ours outperforms recent ``think-with-image'' baselines such as DeepEyes and Thyme. We attribute this advantage to the high-quality training data constructed by our data engine, which specifically targets the unique characteristics of scientific charts—such as high information density and \textbf{multi-subplot layouts}. Unlike generic reasoning models that treat charts as standard images, \ours benefits from fine-grained instruction tuning that explicitly models the structural relationships within complex subplots, leading to more precise visual grounding and reasoning.

\subsection{Results on Out-of-Domain Visual Reasoning Benchmarks}
\label{sec:ood-result}

To verify the generalization ability of \ours beyond the chart domain, we additionally evaluate our model on three general visual reasoning benchmarks: MathVista~\cite{mathvista}, WeMath~\cite{wemath}, and MathVerse~\cite{mathverse}. These benchmarks involve visual understanding combined with multi-step numerical and logical reasoning, and are not limited to chart-based inputs.

\begin{table}[t]
\caption{Results on out-of-domain visual reasoning benchmarks.}
\label{tab:ood-result}
\centering
\small
\setlength{\tabcolsep}{10pt} %
\renewcommand{\arraystretch}{1.0}
\begin{tabular}{l c c c}
\toprule
\textbf{Method} & \textbf{MathVista} & \textbf{WeMath} & \textbf{MathVerse} \\
\midrule

Qwen2.5-VL-3B & \textbf{62.30} & 49.77 & 34.30 \\
\rowcolor{rowpurple} \textbf{\ours-3B} & 60.50 & \textbf{58.45} & \textbf{37.80} \\
 $\Delta$ (\textit{vs} Base) & -1.80 & \dplus{8.68} & \dplus{3.50} \\

\midrule

 Qwen2.5-VL-7B & 68.20 & 64.94 & 45.50 \\
\rowcolor{rowpurple} \textbf{\ours-7B} & \textbf{70.10} & \textbf{67.13} & \textbf{47.10} \\
 $\Delta$ (\textit{vs} Base) & \dplus{1.90} & \dplus{2.19} & \dplus{1.60} \\

\bottomrule
\end{tabular}
\end{table}

As shown in \Cref{tab:ood-result}, despite not being trained on these tasks, \ours shows performance gains across multiple benchmarks. These results indicate that \ours's tool-integrated reasoning framework enables explicit numerical computation and fine-grained visual perception, which in turn generalize effectively to visual mathematical reasoning. By invoking code execution for structured computation, \ours demonstrates strong generalization to out-of-domain visual reasoning benchmarks.

\subsection{Ablation Studies}
\label{sec:ablation}
In this section, we perform ablation studies on several key components in \ours.

\begin{table}[t]
\caption{Ablation study on the effectiveness of tool-integrated reinforcement learning.}\label{tab:ablation-study}
\centering
\small
\setlength{\tabcolsep}{5.5pt} %
\renewcommand{\arraystretch}{1.0}
\begin{tabular}{l c c c c c}
\toprule
\multirow{2.5}{*}{\textbf{Model}} & \multicolumn{2}{c}{\textbf{CharXiv}} & \textbf{ChartQAPro} & \multicolumn{2}{c}{\textbf{ReachQA}} \\
\cmidrule(lr){2-3} \cmidrule(lr){4-4} \cmidrule(lr){5-6}
 & \textit{Reas.} & \textit{Desc.} & \textit{Overall} & \textit{Reas.} & \textit{Recog.} \\
\midrule
Qwen2.5-VL-3B & 31.30 & 58.60 & 30.28 & 26.80 & 59.20 \\
\quad + CoT prompt & 32.40 & 52.00 & 30.64 & 29.70 & 53.30 \\
\quad + Text-only RL & 36.00 & 60.38 & 30.69 & 26.70 & 61.60 \\
\rowcolor{rowpurple} \quad \textbf{+ \ours} & \textbf{41.40} & \textbf{74.75} & \textbf{42.54} & \textbf{36.70} & \textbf{61.40} \\
\bottomrule
\end{tabular}
\end{table}

\begin{table*}[t]
\caption{Performance comparison using different training data sources.}
\label{tab:data-source-comparison}
\centering
\small
\setlength{\tabcolsep}{9.5pt} %
\renewcommand{\arraystretch}{1.0}
\begin{tabular}{l cc c cc cc c c c}
\toprule
\multirow{2.5}{*}{\textbf{Training Data}} & \multicolumn{2}{c}{\textbf{CharXiv}} & \multirow{2.5}{*}{\textbf{ChartX}} & \multicolumn{2}{c}{\textbf{ChartBench}} & \multicolumn{2}{c}{\textbf{ReachQA}} & \textbf{ChartQA} & \textbf{ChartQAPro} & \multirow{2.5}{*}{\textbf{Avg.}} \\
\cmidrule(lr){2-3} \cmidrule(lr){5-6} \cmidrule(lr){7-8} \cmidrule(lr){9-9} \cmidrule(lr){10-10}
 & \textit{Reas.} & \textit{Recog.} & & \textit{NQA} & \textit{Binary} & \textit{Reas.} & \textit{Recog.} & \textit{Avg.} & \textit{Overall} & \\
\midrule
Qwen2.5-VL-3B & 31.30 & 58.60 & 53.91 & 48.10 & 77.26 & 26.80 & 59.20 & 76.64 & 30.28 & 51.34 \\
ReachQA & 35.00 & 72.10 & 54.51 & 48.24 & 83.81 & \textbf{37.40} & 59.90 & 79.52 & 40.83 & 56.81 \\
ECD & 36.00 & 74.40 & 53.82 & 50.43 & 78.17 & 36.00 & 60.20 & 79.80 & 38.97 & 56.42 \\
\rowcolor{rowpurple} \textbf{\data} & \textbf{41.40} & \textbf{74.75} & \textbf{55.38} & \textbf{51.71} & \textbf{84.51} & 36.70 & \textbf{61.40} & \textbf{81.68} & \textbf{42.54} & \textbf{58.90} \\
\bottomrule
\end{tabular}
\end{table*}

\paragraph{Effect of Tool-Integrated Reinforcement Learning.}
\Cref{tab:ablation-study} evaluates Qwen2.5-VL-3B with different reasoning augmentation methods. Chain-of-Thought prompting alone fails to improve performance and even degrades descriptive accuracy, indicating that pure prompt engineering cannot equip the model with sufficient capacity for chart understanding.
Text-only reinforcement learning yields only marginal gains, as language-level optimization cannot resolve errors in fine-grained perception and numerical computation. In contrast, \ours's tool-integrated reasoning leads to substantial improvements across all metrics, with especially large gains on reasoning-intensive benchmarks such as CharXiv and ChartQAPro. These results show that explicit tool invocation is essential for robust chart reasoning.

\paragraph{Comparison with Other Synthesized Training Data.}
To evaluate the effectiveness of our synthesized training data \data, we train \ours on ReachQA~\cite{reachqa} and ECD~\cite{ECD} using the same training pipeline. As shown in \Cref{tab:data-source-comparison}, training on DuoChart yields superior overall performance to training on ReachQA or ECD. This further supports our results in \Cref{sec:dataquality} and highlights the superior quality and complexity of \data, leading to more robust and generalizable chart reasoning capabilities.

\begin{table}[h]
\caption{Ablation study on the training data source.}
\label{tab:data-ablation}
\centering
\small
\setlength{\tabcolsep}{10pt} %
\renewcommand{\arraystretch}{1.0}
\newcommand{\cmark}{\ding{51}} 

\begin{tabular}{cccccc}
\toprule
\multicolumn{2}{c}{\textbf{Data Source}} & \multicolumn{2}{c}{\textbf{CharXiv}} & \multicolumn{2}{c}{\textbf{ReachQA}} \\
\textit{arXiv} & \textit{LLM-Syn} & \textit{Reas.} & \textit{Desc.} & \textit{Reas.}  & \textit{Recog.} \\
\midrule
\multicolumn{2}{c}{Qwen2.5-VL-3B} & 31.30 & 58.60 & 26.80 & 59.20 \\
\midrule
\cmark & & 37.50 & \textbf{76.05} & 33.50  & 57.40 \\
 & \cmark & 37.80 & 72.85 & 35.70 & 59.50 \\
\rowcolor{rowpurple} \cmark & \cmark & \textbf{41.40} & 74.75 & \textbf{36.70} & \textbf{61.40} \\
\bottomrule
\end{tabular}
\end{table}

\paragraph{Impact of Data Composition.}
We analyze the effect of different training data sources in \Cref{tab:data-ablation}. Training with only real-world arXiv chart data substantially improves descriptive performance on CharXiv, reflecting the benefit of exposure to diverse layouts and visual artifacts in scientific figures. In contrast, incorporating code-synthesized data is crucial for reasoning, as programmatically generated questions involving trend analysis and exact numerical comparison. Combining both data sources enables \ours to jointly learn robust visual generalization and precise computational reasoning.

\paragraph{Effect of Different Tools.}
We analyze the contributions of the two primary tools used in \ours: image cropping and code computation. As shown in \Cref{tab:tool-ablation}, both tools improve overall performance and exhibit complementary strengths. We compare \ours with single-tool variants (Crop-only and Code-Computation-only) as well as the base model. The results show that using either tool individually yields gains over the baseline, while integrating both tools achieves the best average performance by jointly supporting fine-grained visual perception (Crop) and multi-step logical and numerical reasoning (Code).

\begin{table}[t]
\caption{Tool usage ablation.}
\label{tab:tool-ablation}
\centering
\small
\setlength{\tabcolsep}{6pt}
\renewcommand{\arraystretch}{1.0}
\newcommand{\cmark}{\ding{51}}

\begin{tabular}{cccccc}
\toprule
\multicolumn{2}{c}{\textbf{Tool Usage}} & \textbf{ChartBench} & \textbf{ReachQA} & \textbf{CharXiv} & \textbf{Avg.} \\
\textit{Crop} & \textit{Code} & & & & \\
\midrule
\multicolumn{2}{c}{Qwen2.5-VL-3B} & 48.10 & 26.80 & 31.30 & 35.40 \\
\midrule
\cmark &  & \textbf{51.95} & 32.20 & 37.90 & 40.68 \\
 & \cmark & 45.52 & 35.60 & 40.20 & 40.44 \\
\rowcolor{rowpurple} \cmark & \cmark & 51.71 & \textbf{36.70} & \textbf{41.40} & \textbf{43.27} \\
\bottomrule
\end{tabular}
\end{table}

\begin{figure}
    \centering
    \includegraphics[width=\linewidth]{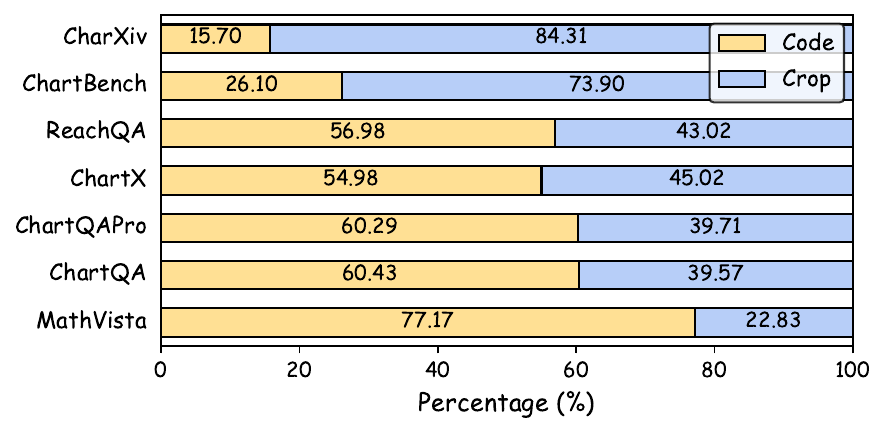}
    \caption{Distribution of tool calls under different benchmarks.}
    \label{fig:tool-distribution}
\end{figure}

\paragraph{Dynamic Tool Selection Strategy.}
We analyze tool call distributions to characterize the dynamic tool selection behavior of \ours. As shown in \Cref{fig:tool-distribution}, CharXiv features more complex layouts and therefore predominantly triggers Crop tool calls. In contrast, benchmarks with simpler layouts but higher numerical complexity rely more on computation tool calls, which constitute 60.43\% of actions on ChartQA and increase to 77.17\% on MathVista. These results indicate that \ours dynamically adapts tool usage to the structural complexity of the task.

\begin{figure*}[!t]
    \centering
    \includegraphics[width=\textwidth]{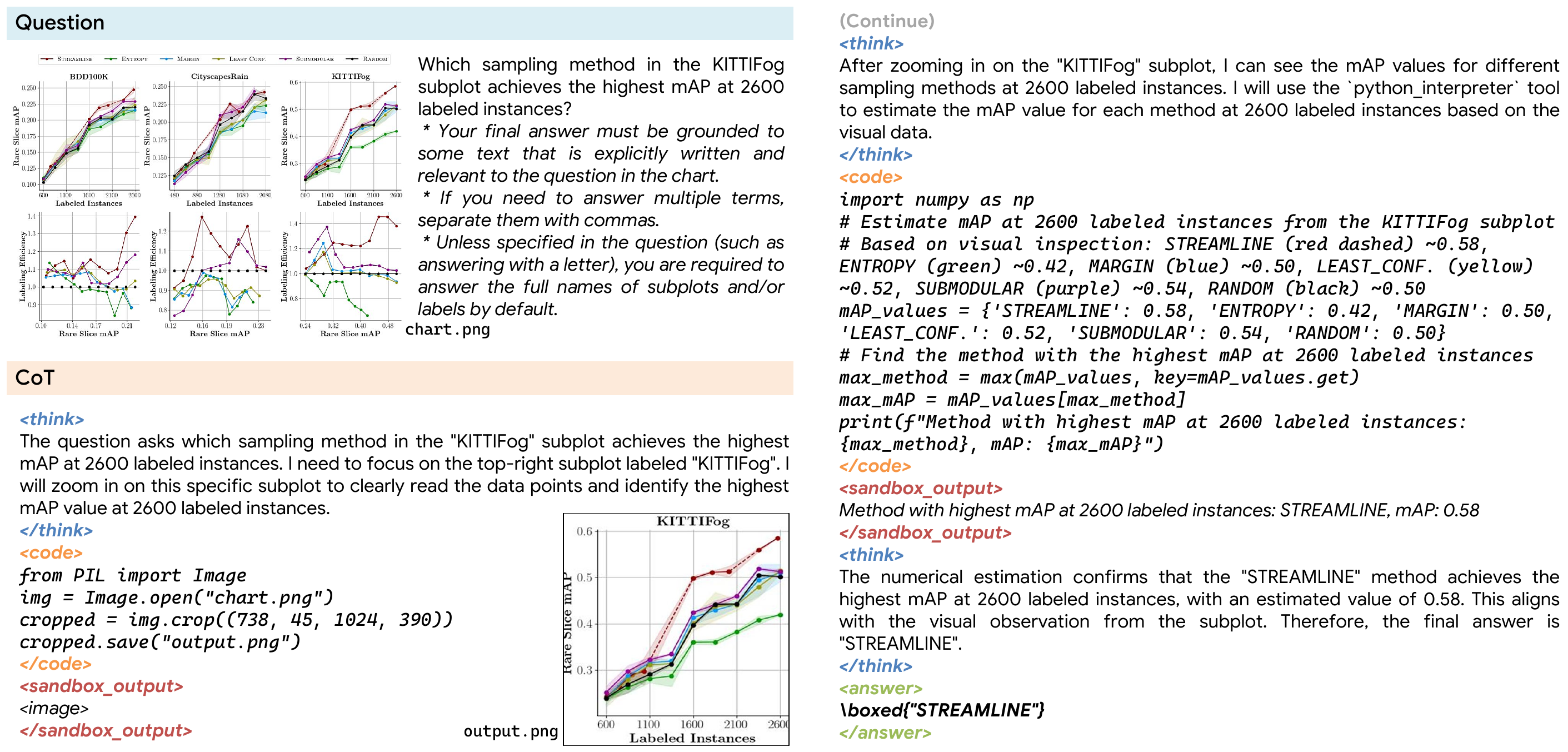}
    \caption{A case study example of \ours on the CharXiv~\cite{charxiv} benchmark.}
    \label{fig:case-study}
\end{figure*}

\subsection{Case Study}

\Cref{fig:case-study} presents a case study of \ours on CharXiv~\cite{charxiv}, a challenging real-world chart benchmark. The example involves a multi-subplot question. \ours first leverages the image-cropping tool to focus on the relevant subplot region, grounding its reasoning in direct visual evidence, and then invokes code-based computation to produce accurate and reliable answers. This structured process demonstrates that \ours goes beyond simple pattern matching by adopting an agentic, step-by-step reasoning workflow. The model exhibits distinct and interpretable reasoning behaviors that closely resemble human cognitive processes. More examples of \ours's tool-integrated reasoning are provided in the supplementary material.

\section{Related Works}

\paragraph{Chart Reasoning.}
Chart reasoning is typically formulated as question answering over charts, requiring both visual understanding and numerical reasoning. Early studies~\citep{pix2struct, deplot, matcha, onechart, chartx} converted charts into textual representations to exploit the reasoning ability of LLMs, but such modality conversion leads to information loss. Subsequent works~\citep{chartllama, chartadapter, chart-based-reasoning, mmc, chartassistant, ureader, mplug-owi2} adopted end-to-end instruction tuning for MLLMs to directly comprehend charts. More recent studies~\citep{chartthinker, domino, synthesizesbs, tinychart} introduced Chain-of-Thoughts or Program-of-Thoughts strategies to enhance reasoning, yet they still mainly rely on internal knowledge of MLLMs. In contrast, \ours integrates MLLMs with external tools, such as code execution for precise numerical computation and localized visual operations~(e.g., cropping), to further strengthen reasoning accuracy and visual perception in chart reasoning.

\paragraph{Tool-Integrated Reasoning.}
Many studies~\citep{toolrl,searchr1,r1searcher,torl,ma2026empowering} on text-based LLMs have demonstrated that incorporating external tools can significantly enhance the reasoning ability of models, giving rise to the paradigm of Tool-Integrated Reasoning~(TIR). Recent MLLM works~\citep{deepeyes, deepeyesv2, vtoolr1, minio3, grit, VisualAR, vlmr3, adaptiveCR} further adopt TIR and employ reinforcement learning techniques to improve multimodal reasoning performance. However, most of these works focus on natural images, where reasoning mainly involves semantic understanding and high-level perception. In contrast, chart reasoning requires finer-grained analysis, such as precise numerical computation and trend interpretation, which poses greater challenges. To address this, we construct chart-specific data for TIR and enable the model to better handle complex reasoning.

\paragraph{Data Synthesis for MLLM Training.} Training MLLMs requires paired image-text data. Some works~\citep{provision, lamm, fm2ds, dspt, IEA} synthesize instructions based on existing image datasets, while others directly generate paired images and text, aiming to circumvent the need for large-scale image collection. Unlike natural images, charts can be losslessly generated from structured markups or programming code, which provides an opportunity to leverage code-capable LLMs for chart synthesis~\cite{matplotagent, chartmimic, ECD, reachqa}. However, current code-driven synthesis pipelines often suffer from limited diversity and suboptimal visual quality. To overcome these limitations, we propose a hybrid framework that integrates code-driven synthesis with real-world chart collection to produce high-quality chart data.

\section{Conclusion}

In this paper, we introduce \data and \ours for chart understanding. We design a multi-agent data synthesis pipeline with rigorous verification and filtering to produce a diverse and challenging dataset that overcomes the limitations of existing synthetic data. By integrating tool-based reasoning with image cropping and code computation, \ours supports both fine-grained visual perception and precise numerical reasoning. These capabilities are essential for interpreting complex real-world charts. Extensive experiments demonstrate that \ours consistently improves performance across chart benchmarks and exhibits strong generalization to out-of-domain visual math reasoning tasks.

\section*{Acknowledgments}
This work was supported by the China NSFC Project (No. 92370206, No. 62576212, No. 62120106006, No. U23B2057) and the Key Research and Development Program of Jiangsu Province, China (No. BF2025029).

\bibliographystyle{ACM-Reference-Format}
\balance
\bibliography{acmmm}

\newpage
\appendix
\onecolumn

\section{Limitations and Impacts}
While our approach achieves substantial improvements across diverse chart reasoning benchmarks, several limitations point to promising directions for future work. First, although our framework currently focuses on two core tools, visual cropping and code computation, these tools are unified under a single code execution interface. This design makes it straightforward to extend our approach to support additional vision or reasoning tools in the future. Second, this work focuses on chart understanding, and extending the methodology to broader visual reasoning domains, such as tables, manuals, and geometric figures, remains an important direction. Third, more fine-grained, process-level reward signals in agentic reinforcement learning may further improve tool usage and multi-step reasoning behaviors.

Despite these limitations, \ours introduces a high-quality chart dataset and a tool-integrated reasoning framework that significantly advances agentic chart understanding. By grounding reasoning in explicit visual operations and computation, our approach enables more reliable and robust chart analysis, with potential impact on scientific research, financial analysis, and other data-driven decision-making scenarios. We hope this work inspires further exploration of tool-integrated reasoning for chart understanding.

\section{Training Details}
\label{app:train-detail}

\subsection{Cold Start SFT Details}
We utilize the Qwen2.5-VL series~\cite{qwen25vl} as our base models. During the Cold-Start SFT stage, we fine-tune the models to establish basic instruction-following and tool-usage capabilities. The detailed hyperparameters employed for both the 3B and 7B model variants are summarized in \Cref{tab:sft_hyperparameters}. The cold-start SFT is implemented using SWIFT~\cite{swift} framework.

\begin{table}[h]
    \centering    
    \small
    \caption{Detailed hyperparameter configuration for the Cold-Start SFT stage.}
    \label{tab:sft_hyperparameters}
    \renewcommand{\arraystretch}{1.1}
    \begin{tabular}{lcc}
        \toprule
        \textbf{Hyperparameter} & \textbf{3B} & \textbf{7B} \\
        \midrule
        \texttt{num\_train\_epochs} & 3 & 3 \\
        \texttt{train\_batch\_size} & 64 & 64 \\
        \texttt{per\_device\_train\_batch\_size} & 2 & 1 \\
        \texttt{gradient\_accumulation\_steps} & 4 & 8 \\
        \texttt{learning\_rate} & 1e-5 & 1e-5 \\
        \texttt{warmup\_ratio} & 0.05 & 0.05 \\
        \texttt{max\_length} & 32768 & 32768 \\
        \bottomrule
    \end{tabular}
\end{table}

\subsection{Reinforcement Learning Details}
We perform reinforcement learning starting from the cold-start model, utilizing the widely adopted VeRL~\cite{verl} framework. During the RL stage, we train on \data, excluding all samples used for cold-start to avoid data leakage. Because our training data primarily consists of open-ended text-answer questions (\Cref{fig:data-stat}), exact matching can lead to inaccurate rewarding. To address this, we employ an LLM-as-a-Judge approach~\cite{survey-llm-judge, generation-llm-judge} to evaluate answer accuracy. Specifically, we use CompassVerifier-7B~\cite{compassverifier}, an efficient and reliable judge model that outperforms GPT-4o~\cite{gpt4o} and DeepSeek-V3~\cite{deepseekv3}, to compare model predictions with ground-truth answers.

\paragraph{Tool Execution Environment.}
We unify all external tools through a code execution interface. Specifically, we follow ToRL~\cite{torl} and adopt SandboxFusion~\cite{sandbox} as the execution backend to safely run generated code within isolated sandboxed environments. SandboxFusion supports asynchronous and parallel code execution, which allows us to efficiently perform large-scale rollouts during reinforcement learning. This environment ensures both execution safety and high throughput, enabling stable and scalable RL training.

\paragraph{Training Hyperparameters} We employ Group Relative Policy Optimization (GRPO)~\cite{grpo} for the RL training stage. \Cref{tab:grpo_hyperparameters} presents the key training configurations. Unless explicitly listed, all other hyperparameters follow standard default practices.

\begin{table}[h]
    \centering
    \small
    \caption{Hyperparameters used for GRPO RL training. We report parameters for both model scales.}
    \label{tab:grpo_hyperparameters}    
    \renewcommand{\arraystretch}{1.1}
    \begin{tabular}{lcc}
        \toprule
        \textbf{Hyperparameter} & \textbf{3B} & \textbf{7B} \\
        \midrule
        \texttt{trainer.total\_epochs} & 1 & 1 \\
        \texttt{data.train\_batch\_size} & 128 & 128 \\
        \texttt{actor\_rollout\_ref.actor.ppo\_mini\_batch\_size} & 128 & 128 \\
        \texttt{actor\_rollout\_ref.actor.optim.lr} & 1e-6 & 1e-6 \\
        \texttt{data.max\_prompt\_length} & 8192 & 8192 \\
        \texttt{data.max\_response\_length} & 16384 & 16384 \\
        \texttt{actor\_rollout\_ref.actor.kl\_loss\_coef} & 0.001 & 0.001 \\
        \texttt{actor\_rollout\_ref.actor.entropy\_coeff} & 0 & 0 \\
        \texttt{actor\_rollout\_ref.rollout.n} & 8 & 8 \\
        \texttt{actor\_rollout\_ref.rollout.temperature} & 1.0 & 1.0 \\
        \texttt{actor\_rollout\_ref.rollout.max\_num\_batched\_tokens} & 32768 & 32768 \\
        \texttt{actor\_rollout\_ref.rollout.multi\_turn.max\_user\_turns} & 5 & 4 \\
        \texttt{actor\_rollout\_ref.rollout.multi\_turn.max\_assistant\_turns} & 5 & 4\\
        \bottomrule
    \end{tabular}
\end{table}

\section{Evaluation Details}
\label{app:eval-detail}
\subsection{Dataset Quality Assessment}
Following ECD~\cite{ECD}, we assessed the complexity of chart image through the \textbf{Average Pixel Entropy Criteria}. Concretely, the average pixel entropy of the whole Dataset $\mathcal{D}= \{ x_1\ldots x_N\}$ can be defined as:
\begin{equation}
    \mathcal{H}_{\text{avg}}(\mathcal{D}) = - \frac{1}{N} \sum_{j=1}^{N} \sum_{i=1}^{L} p(i | x_j) \log p(i | x_j)
\end{equation}
where $p(i|x_j)$ is the normalized histogram probability of intensity value $i$, $L$ is the number of discrete intensity levels (\textit{e.g.}, 256 for 8-bit images), and $N$ is the size of Dataset $\mathcal{D}$.

We also employed GPT-5.2 to conduct a comprehensive evaluation, scoring all metrics on a scale of 1 to 5. Specifically, we randomly sampled 500 chart images from each dataset for the model to assess \textbf{Visual Quality}. For questions-related metrics, we selected 500 recognition and 500 reasoning questions randomly and prompted model to assess \textbf{QA-alignment} and \textbf{Answer Correctness}, calculating the average scores. Finally, \textbf{Reasoning Difficulty} was evaluated solely on the reasoning subset.

\subsection{Baseline Details}
We evaluate all models on the benchmarks concerning chart reasoning, following each benchmark's official protocol and reporting accuracy as the primary matric. For proprietary models, we evaluate them by accessing their official APIs. For general-purpose, chart-specialized, and think-with-image models, we load the released checkpoints from HuggingFace\footnote{\url{https://huggingface.co/}} and follow the official inference code provided in their repositories to ensure a reproducible and fair evaluation environment.

\subsection{Benchmark Details}
We evaluated \ours across a suite of benchmarks to assess its performance in understanding and reasoning of charts, using charts derived from both the real world and the synthetic pipeline. We also assessed the out-of-domain generality of our method on three visual-mathematical reasoning benchmarks. The details of these benchmarks are listed below.

For real-word chart data, \textbf{CharXiv}~\cite{charxiv} benchmark contains 2,323 challenging charts collected from arXiv academic papers. We used the official validation split that includes 4,000 descriptive questions and 1,000 reasoning questions for evaluation. \textbf{ChartQA}~\cite{chartqa} benchmark comprises charts crawled from four different sources and questions acquired from human-generated and augmented segments. We used the official test set consisting of 1,509 images and the corresponding 2,500 QA pairs. \textbf{ChartQAPro}~\cite{chartqapro} benchmark includes 1,341 chart images from 157 online platforms and 1948 QA pairs annotated by human-VLM collaboration processes. We used the whole dataset for evaluation.

Among the synthetic benchmarks, \textbf{ChartBench}~\cite{chartbench} benchmark consists of 2,100 chart images paired with 18,900 questions (2,100 numerical questions (NQA) and 16,800 binary ``yes/no''). Each chart data within \textbf{ChartX}~\cite{chartx} benchmark includes four modalities such as image, Comma-Separated Values (CSV), python code and text description. We only used the images and the corresponding 1,152 QA pairs in the official test set for evaluation. For \textbf{ReachQA}~\cite{reachqa}, we used the test split containing 500 chart images and 2,000 questions (1,000 recognition-oriented and 1,000 reasoning-oriented questions).

Finally, we introduced three visual-mathematical reasoning benchmarks to evaluate the generality of our method. \textbf{MathVista}~\cite{mathvista} is a benchmark designed to combine difficulty from diverse mathematical and visual tasks, challenging the model's fine-grained perception and compositional reasoning capabilities. We used the official testmini split which includes 1,000 images and 1,000 QA pairs. \textbf{WeMath}~\cite{wemath} is a visual mathematical benchmark spaning hierarchical knowledge concepts and knowledge granularity. We used the official testmini subset containing 1,740 examples for evaluation. \textbf{MathVerse}~\cite{mathverse} is an all-around visual math benchmark designed for an equitable and in-depth evaluation of MLLMs. We evaluated our method using the testmini split consisting of 2,364 images and 3,940 QA pairs.

We evaluate all models using the default prompts provided by each benchmark and set the sampling temperature to 0 to ensure deterministic generation. For \ours, we deploy the same code sandbox environment as used during the reinforcement learning stage.

\section{More Results}
\label{app:more-results}

\begin{table}[h]
\caption{Effect of different training stages. Results are reported for the baseline, cold-start SFT, and \ours at both 3B and 7B model scales.}
\label{tab:sft-ablation}
\centering
\small
\setlength{\tabcolsep}{5pt} %
\renewcommand{\arraystretch}{1.1}
\begin{tabular}{l cc c cc c}
\toprule
\multirow{2.5}{*}{\textbf{Model}} & \multicolumn{2}{c}{\textbf{CharXiv}} & \textbf{ChartQAPro} & \multicolumn{2}{c}{\textbf{ReachQA}} & \multirow{2.5}{*}{\textbf{Avg.}} \\
\cmidrule(lr){2-3} \cmidrule(lr){4-4} \cmidrule(lr){5-6}
 & \textit{Reas.} & \textit{Desc.} & \textit{Overall} & \textit{Reas.} & \textit{Recog.} & \\
\midrule
\multicolumn{7}{l}{\textit{\textbf{3B Scale}}} \\
Qwen2.5-VL-3B & 31.30 & 58.60 & 30.28 & 26.80 & 59.20 & 41.24 \\
Cold-start SFT & 33.80 & 72.10 & 41.64 & 33.40 & 55.20 & 47.23 \\
\rowcolor{rowpurple} \textbf{\ours-3B} & \textbf{41.40} & \textbf{74.75} & \textbf{42.54} & \textbf{36.70} & \textbf{61.40} & \textbf{51.36} \\
\midrule
\multicolumn{7}{l}{\textit{\textbf{7B Scale}}} \\
Qwen2.5-VL-7B & 42.90 & 74.15 & 39.84 & 37.70 & 70.50 & 53.02 \\
Cold-start SFT & 44.40 & \textbf{82.00} & 49.24 & 44.70 & 63.20 & 56.71 \\
\rowcolor{rowpurple} \textbf{\ours-7B} & \textbf{50.50} & 80.30 & \textbf{49.62} & \textbf{46.30} & \textbf{72.30} & \textbf{59.80} \\
\bottomrule
\end{tabular}
\end{table}

\paragraph{Effect of training stages.}
As shown in \Cref{tab:sft-ablation}, the two-stage training procedure leads to progressive performance improvements. Cold-start supervised fine-tuning provides a strong initialization by enabling the model to acquire basic tool-use behaviors, while the subsequent reinforcement learning stage further refines tool invocation and multi-step reasoning strategies, resulting in consistent gains across benchmarks.

\begin{table*}[h]
\caption{Distribution of tool usage across different benchmarks. The values represent the percentage (\%) of calls for \textbf{Code Computation} versus \textbf{Image Cropping}. The order is arranged from perception-heavy tasks (left) to computation-heavy tasks (right), illustrating the model's adaptive strategy.}
\label{tab:tool-distribution}
\centering
\small
\setlength{\tabcolsep}{5.0pt} %
\renewcommand{\arraystretch}{1.1}
\begin{tabular}{l cc c c c c c c c c}
\toprule
\multirow{2.5}{*}{\textbf{Tool Type}} & \multicolumn{2}{c}{\textbf{CharXiv}} & \multirow{2.5}{*}{\textbf{ChartQA}} & \textbf{ChartQA} & \textbf{Chart} & \multirow{2.5}{*}{\textbf{ChartX}} & \multirow{2.5}{*}{\textbf{ReachQA}} & \textbf{Math} & \textbf{We} & \textbf{Math} \\
 & \textit{Reas.} & \textit{Desc.} & & \textbf{Pro} & \textbf{Bench} & & & \textbf{Vista} & \textbf{Math} & \textbf{Verse} \\
\midrule
Code Computation & 18.31 & 13.08 & 60.43 & 60.29 & 26.10 & 54.98 & 56.98 & 77.17 & 80.46 & 80.46 \\
Image Crop & 81.69 & 86.92 & 39.57 & 39.71 & 73.90 & 45.02 & 43.02 & 22.83 & 19.54 & 19.54 \\
\bottomrule
\end{tabular}
\end{table*}

\paragraph{Dynamic Tool Selection Strategy.}
Detailed tool usage distribution across all benchmarks, including six chart benchmarks and three out-of-domain visual reasoning tasks, are presented in \Cref{tab:tool-ablation}. We observe that tasks with more complex layouts, such as CharXiv, predominantly trigger the use of visual cropping tools, while benchmarks focused on mathematical reasoning with simpler images, such as MathVista, rely more heavily on computation tools. These findings demonstrate that \ours dynamically adapts its tool selection strategy to the structural and cognitive demands of each task.

\section{Data Statistics}
\label{app:data-stats}

We present detailed data statistics in this section, covering the distribution of chart image sources (\Cref{tab:chart_source_dist}), chart types at both the figure level (\Cref{tab:chart_type_dist_fig}) and subplot level (\Cref{tab:chart_type_dist_subplot}), the distribution of the number of subplots per figure (\Cref{tab:num_chart_dist}), answer types (\Cref{tab:answer_type_dist}), and question types (\Cref{tab:question_type_dist}).

\begin{table}[h]
\centering
\small
\setlength{\tabcolsep}{10pt}
\caption{Distribution of Chart Sources.}
\label{tab:chart_source_dist}
\begin{tabular}{p{0.3\linewidth} r r}
\toprule
\textbf{Source} & \textbf{Share (\%)} & \textbf{Count} \\
\midrule
LLM-code & 52.08 & 8691 \\
arXiv    & 47.92 & 7997 \\
\midrule
\textbf{Total} &  & \textbf{16688} \\
\bottomrule
\end{tabular}
\end{table}

\begin{table}[h]
\centering
\small
\setlength{\tabcolsep}{10pt}
\caption{Distribution of Chart Types (Figure Level).}
\label{tab:chart_type_dist_fig}
\begin{tabular}{p{0.3\linewidth} r r}
\toprule
\textbf{Category} & {\textbf{Share (\%)}} & \textbf{Count} \\
\midrule
\textbf{One-plot} & 51.32 & 8565 \\
\quad Line chart              & 15.17 & 2531 \\
\quad Bar chart               &  6.75 & 1126 \\
\quad Area chart              &  5.13 &  856 \\
\quad Scatter chart           &  5.08 &  848 \\
\quad Box chart               &  4.46 &  744 \\
\quad Radar chart             &  4.22 &  705 \\
\quad Pie chart               &  3.24 &  541 \\
\quad Heatmap                 &  1.77 &  296 \\
\quad Histogram               &  0.63 &  105 \\
\quad Specific chart (Others) &  4.87 &  813 \\
\midrule
\textbf{Multi-subplot} & 48.68 & 8123 \\
\quad Only one type & 37.81 & 6309 \\
\quad Composite     & 10.87 & 1814 \\
\midrule
\textbf{Total} &  & \textbf{16688} \\
\bottomrule
\end{tabular}
\end{table}

\begin{table}[h]
\centering
\small
\setlength{\tabcolsep}{10pt}
\caption{Distribution of the Number of Subplots per Figure.}
\label{tab:num_chart_dist}
\begin{tabular}{p{0.3\linewidth} r r}
\toprule
\textbf{\# Charts} & \textbf{Share (\%)} & \textbf{Count} \\
\midrule
1    & 51.32 & 8565 \\
2--4 & 29.61 & 4941 \\
5--9 & 18.26 & 3047 \\
9+   &  0.81 &  135 \\
\midrule
\textbf{Average \# Subplots}&  & \textbf{2.78} \\
\bottomrule
\end{tabular}
\end{table}

\begin{table}[h]
\centering
\small
\setlength{\tabcolsep}{10pt}
\caption{Distribution of Chart Types (Subplot Level).}
\label{tab:chart_type_dist_subplot}
\begin{tabular}{p{0.3\linewidth} r r}
\toprule
\textbf{Category} & {\textbf{Share (\%)}} & \textbf{Count} \\
\midrule
Line chart      & 30.07 & 13954 \\
Bar chart       & 20.37 &  9451 \\
Scatter chart   & 11.45 &  5312 \\
Heatmap         &  9.14 &  4240 \\
Pie chart       &  8.14 &  3775 \\
Area chart      &  4.59 &  2132 \\
Box chart       &  4.14 &  1922 \\
Histogram       &  2.44 &  1132 \\
Radar chart     &  1.67 &  775 \\
Special charts (Others)   &  8.00 &  3711 \\
\midrule
\textbf{Total} & & \textbf{46404} \\
\bottomrule
\end{tabular}
\end{table}

\begin{table}[h]
\centering
\small
\setlength{\tabcolsep}{10pt}
\caption{Distribution of Answer Types.}
\label{tab:answer_type_dist}
\begin{tabular}{p{0.3\linewidth} r r}
\toprule
\textbf{Answer Type} & \textbf{Share (\%)} & \textbf{Count} \\
\midrule
Text          & 74.26 & 78045 \\
Numeric value & 18.59 & 19533 \\
Binary        &  5.51 &  5791 \\
List / Range  &  1.64 &  1727 \\
\midrule
\textbf{Total} &  & \textbf{105096} \\
\bottomrule
\end{tabular}
\end{table}

\begin{table}[h]
\centering
\small
\setlength{\tabcolsep}{10pt}
\caption{Distribution of Question Types.}
\label{tab:question_type_dist}
\begin{tabular}{p{0.3\linewidth} r r}
\toprule
\textbf{Question Type} & \textbf{Share (\%)} & \textbf{Count} \\
\midrule
\textbf{Recognition} & 57.18 & 60096 \\
\quad Axis label extraction         & 14.32 & 15045 \\
\quad Color extraction              & 13.62 & 14312 \\
\quad Title extraction              &  7.34 &  7716 \\
\quad Tick extraction               &  5.58 &  5869 \\
\quad Numerical value extraction    &  5.56 &  5847 \\
\quad Counting                      &  5.94 &  6246 \\
\quad Pattern Recognition           &  4.78 &  5027 \\
\quad Enumeration                   &  0.03 &    34 \\
\midrule
\textbf{Reasoning} & 42.82 & 45000 \\
\quad Extreme Value Analysis                    & 15.46 & 16249 \\
\quad Conditional Reasoning                     &  9.34 &  9813 \\
\quad Comparative Analysis                      &  6.01 &  6316 \\
\quad Trend Analysis                            &  3.32 &  3484 \\
\quad Aggregation \& Calculation                &  3.20 &  3366 \\
\quad Ranking \& Ordering                       &  2.22 &  2329 \\
\quad Proportional \& Distributional Analysis   &  1.78 &  1874 \\
\quad Pattern \& Correlation                    &  1.49 &  1569 \\
\midrule
\textbf{Total} &  & \textbf{105096} \\
\bottomrule
\end{tabular}
\end{table}

\clearpage

\section{Case Studies}
\label{app:case-study}

This section provides case studies demonstrating how our tool-integrated reasoning model operates on a variety of reasoning tasks.

\begin{figure}[!h]
    \centering
    \includegraphics[width=0.8\linewidth,trim=1000 700 1100 1100,clip]{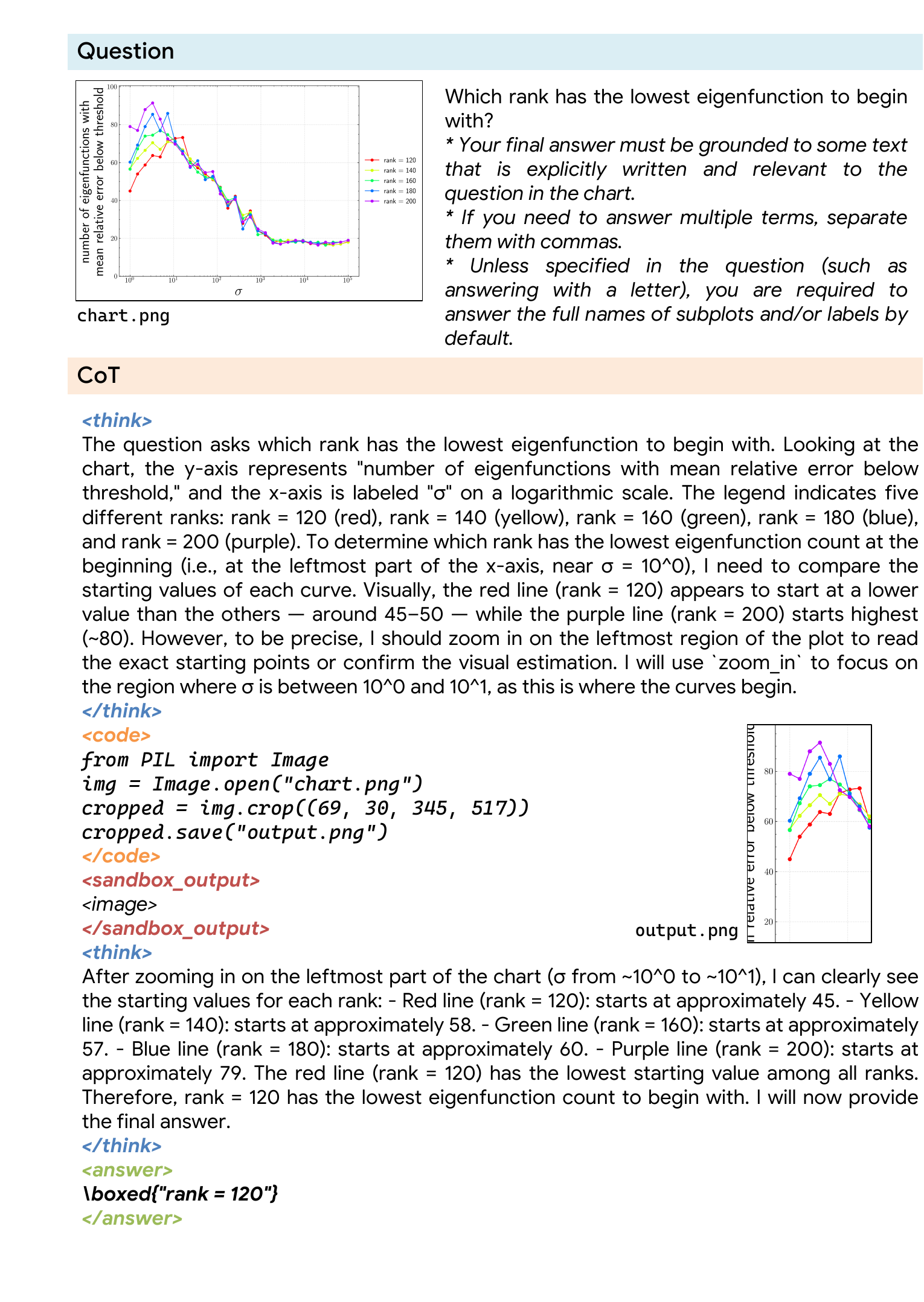}
    \caption{A \ours example using Crop Tool.}
    \label{fig:example1}
\end{figure}

\begin{figure}[!t]
    \centering
    \includegraphics[width=0.8\linewidth,trim=1000 700 1100 1100,clip]{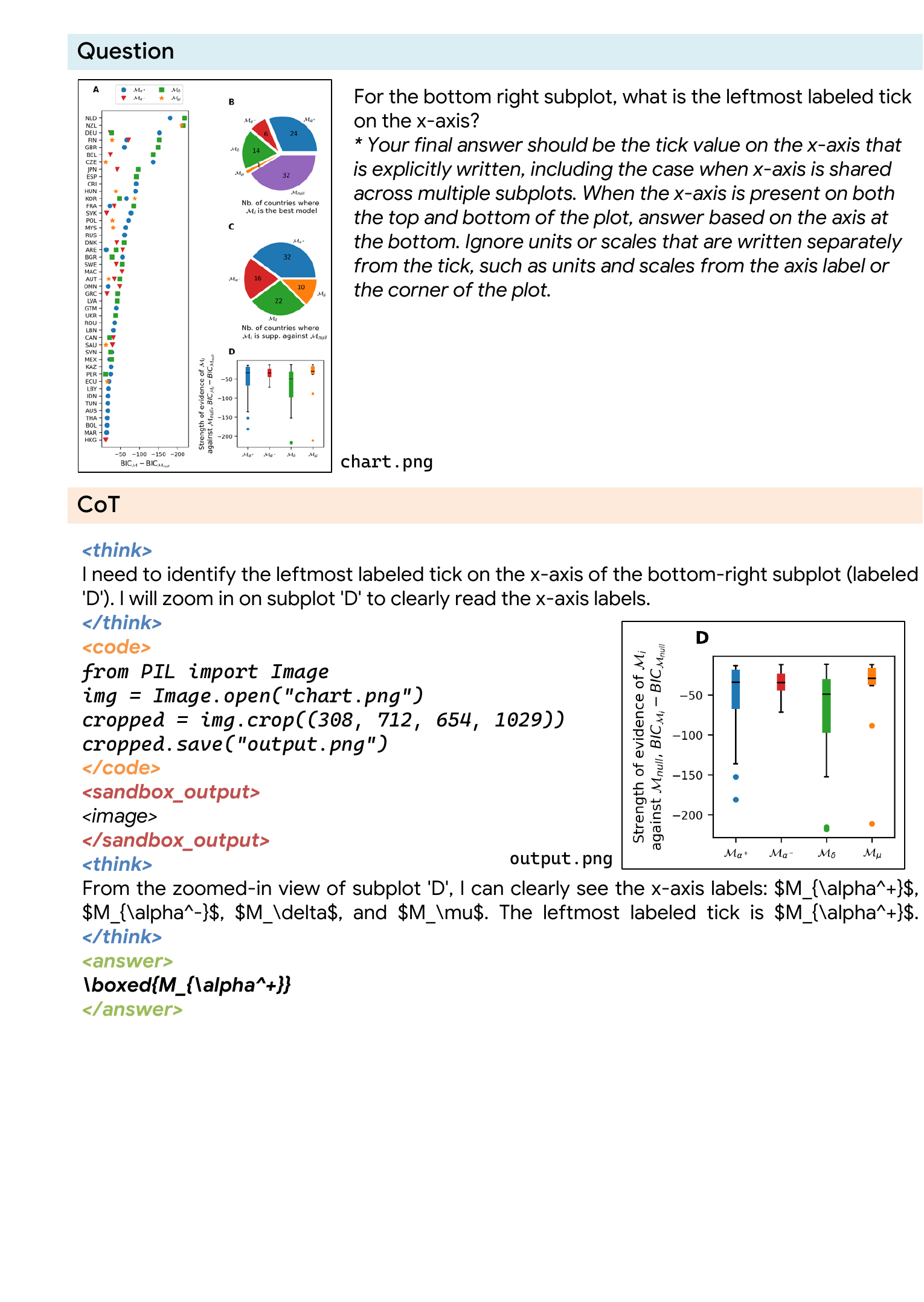}
    \caption{A \ours example using Crop Tool.}
    \label{fig:example2}
\end{figure}

\begin{figure}[!t]
    \centering
    \includegraphics[width=0.8\linewidth,trim=1000 700 1100 1100,clip]{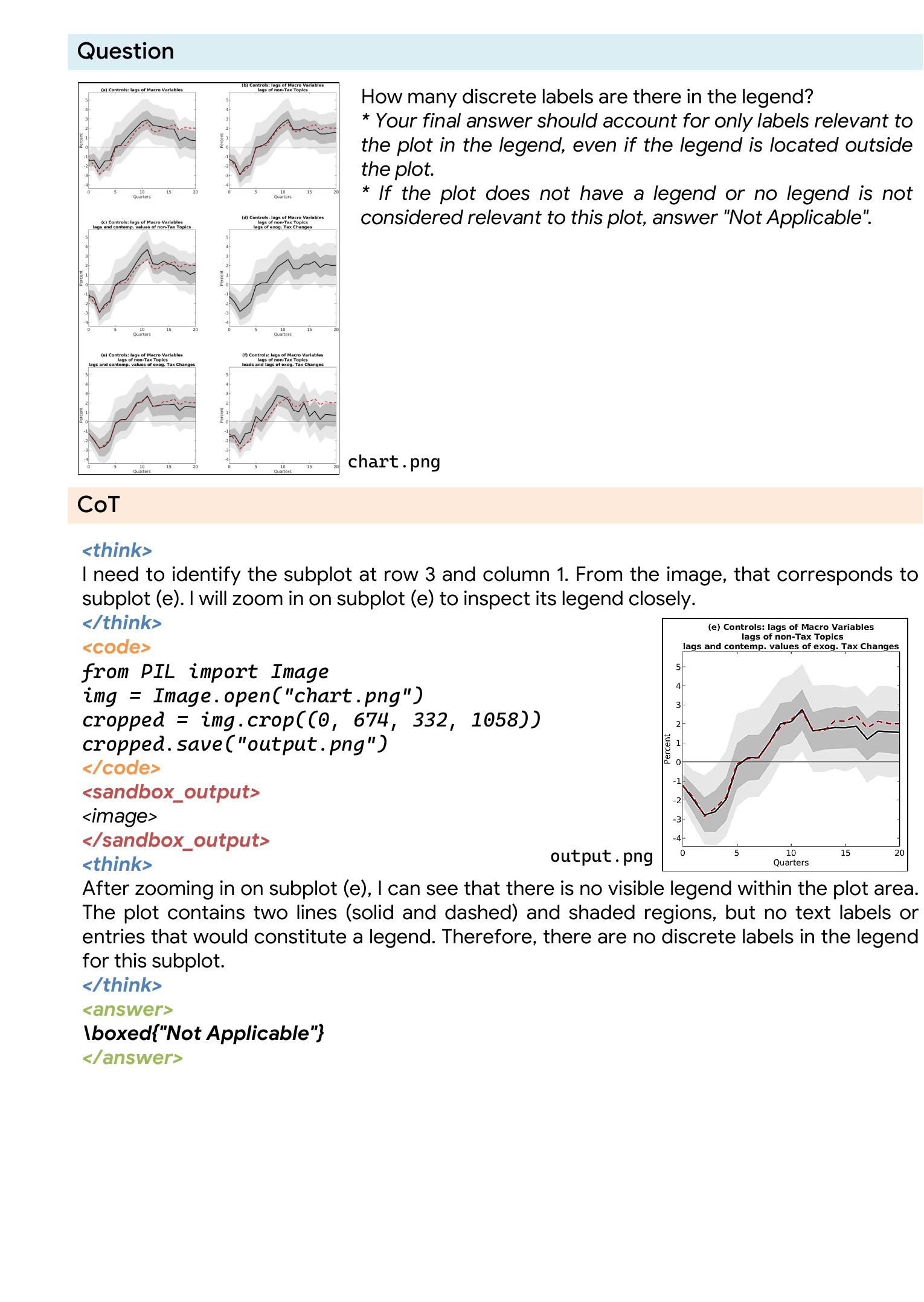}
    \caption{A \ours example using Crop Tool.}
    \label{fig:example3}
\end{figure}

\begin{figure}[!t]
    \centering
    \includegraphics[width=0.8\linewidth,trim=1000 700 1100 1100,clip]{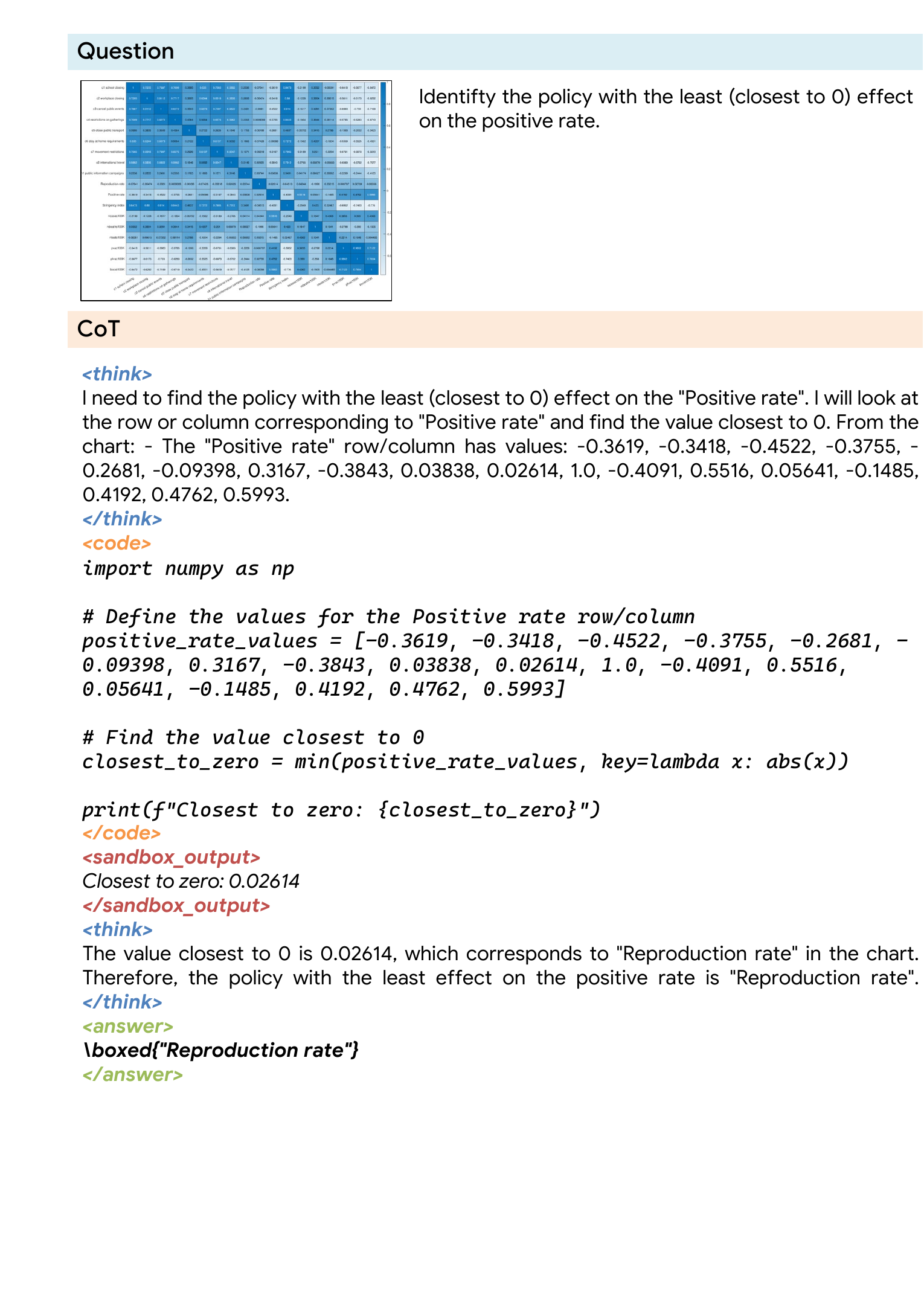}
    \caption{A \ours example using Code Computation Tool.}
    \label{fig:example4}
\end{figure}

\begin{figure}[!t]
    \centering
    \includegraphics[width=0.8\linewidth,trim=1000 700 1100 1100,clip]{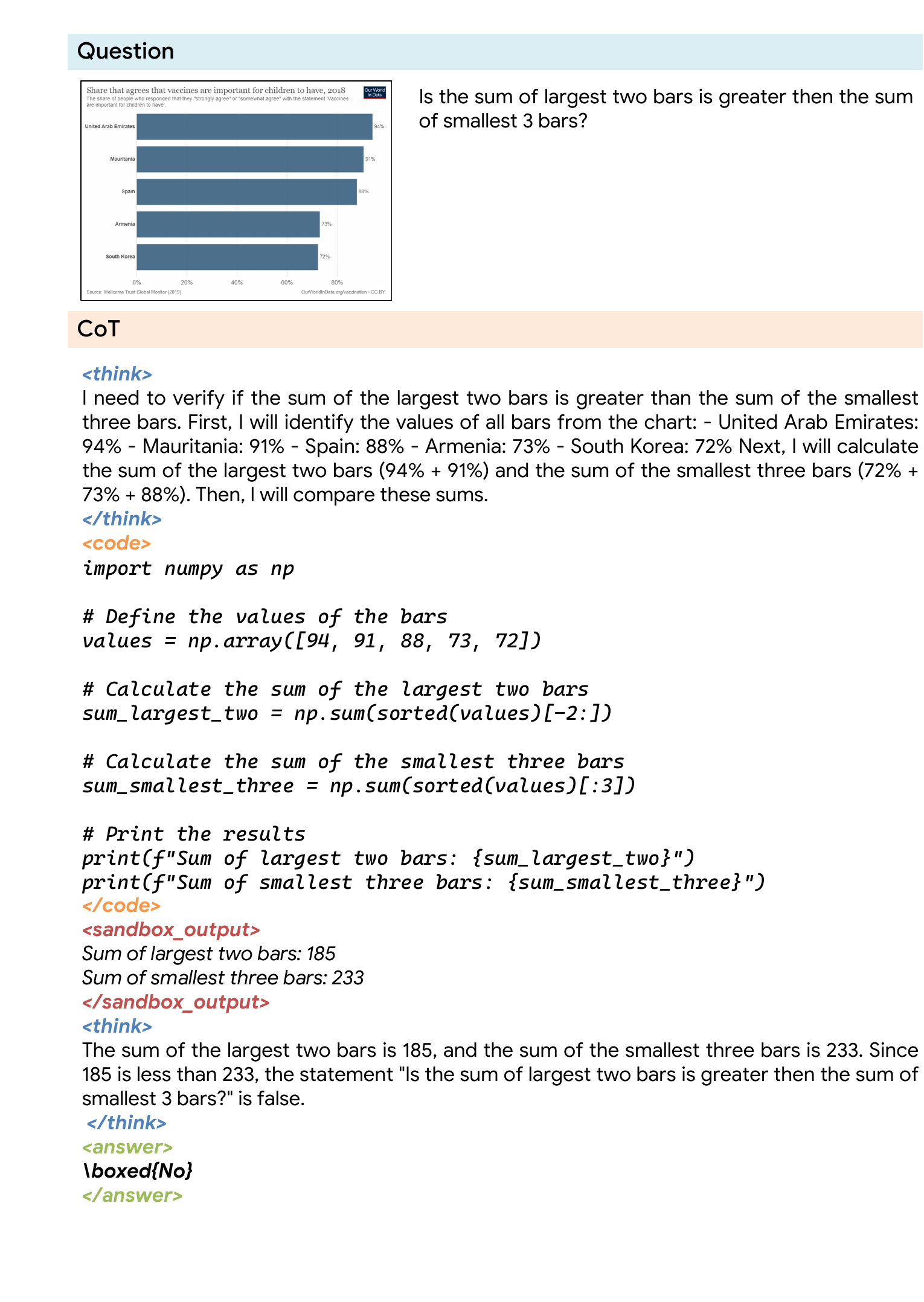}
    \caption{A \ours example using Code Computation Tool.}
    \label{fig:example5}
\end{figure}

\begin{figure}[!t]
    \centering
    \includegraphics[width=0.8\linewidth,trim=1000 700 1100 1100,clip]{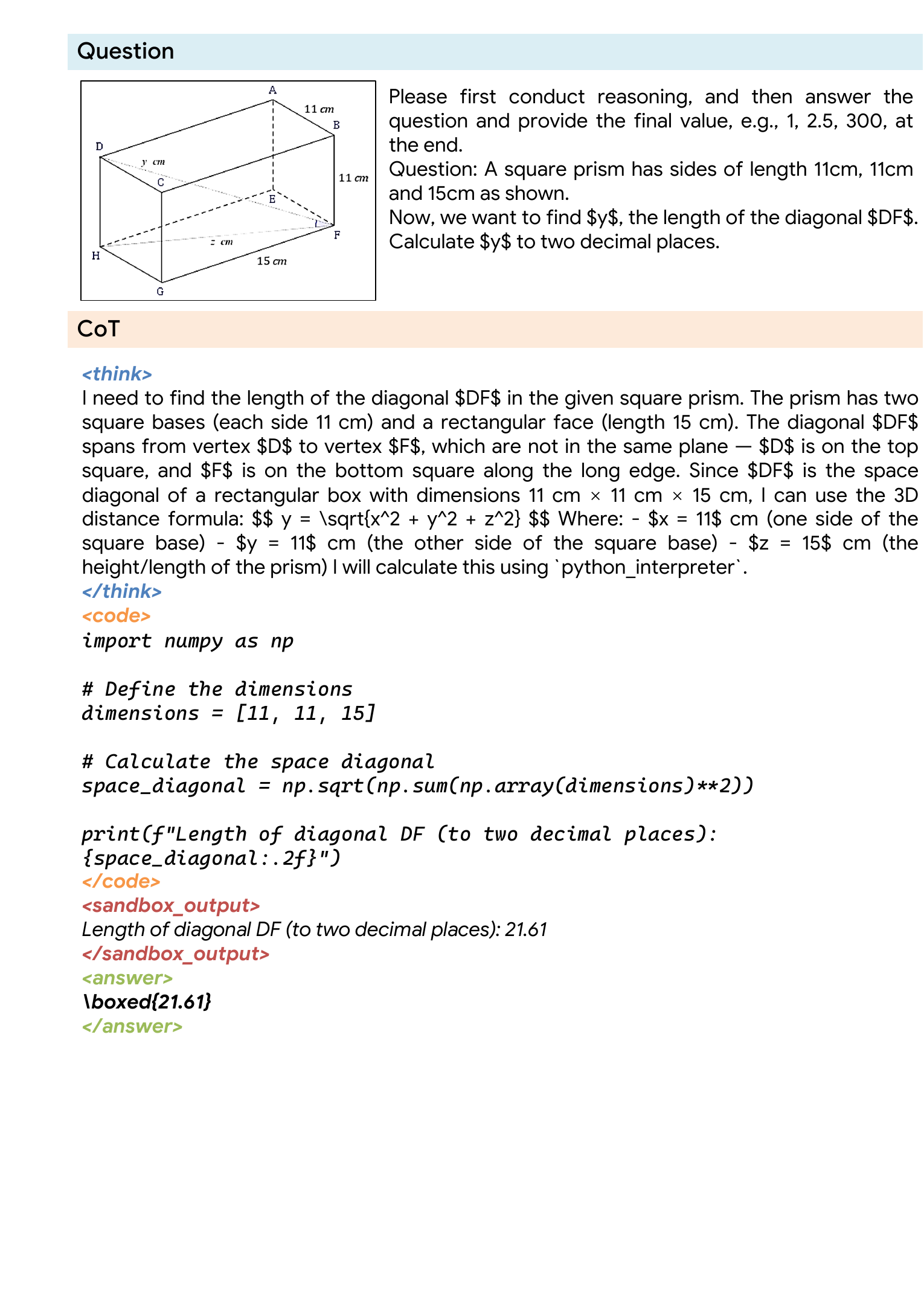}
    \caption{A \ours example using Code Computation Tool on MathVerse~\cite{mathverse} benchmark.}
    \label{fig:example6}
\end{figure}

\end{document}